\documentclass[10pt]{article}

\usepackage{tgpagella}      
\usepackage{mathpazo}       
\usepackage{inconsolata}    

\usepackage[letterpaper, margin=0.75in, top=0.6in, bottom=0.7in]{geometry}

\usepackage[T1]{fontenc}
\usepackage[utf8]{inputenc}
\usepackage{microtype}
\usepackage{parskip}        
\usepackage{graphicx}
\usepackage{subcaption}
\usepackage[dvipsnames,table]{xcolor}
\usepackage{array}
\usepackage{booktabs}
\usepackage{amsmath}
\usepackage{amssymb}
\usepackage{amsthm}
\usepackage{amsfonts}
\usepackage{nicefrac}
\usepackage{algorithm}
\usepackage{algpseudocode}
\usepackage{wrapfig}
\definecolor{linkblue}{HTML}{000080}
\usepackage[
    colorlinks=true,
    linkcolor=linkblue,
    citecolor=linkblue,
    urlcolor=linkblue
  ]{hyperref}
\usepackage{url}
\usepackage{natbib}
\setcitestyle{authoryear,round,citesep={;},aysep={,},yysep={;}}
\usepackage{caption}
\usepackage{enumitem}
\setlist{leftmargin=*}
\usepackage{fontawesome5}

\definecolor{darkred}{rgb}{0.6, 0, 0}
\definecolor{echoline}{HTML}{DB2777}
\definecolor{echobg}{HTML}{FCE7F3}
\definecolor{rowbase}{HTML}{F4F4F5}
\definecolor{rowrl}{HTML}{E6F4F2}
\definecolor{rowecho}{HTML}{FCE7F3}

\graphicspath{{../}}

\begin{document}

\begin{center}
{\LARGE\bf ECHO: Terminal Agents Learn World Models for Free}\\[8pt]
{\normalsize Vaishnavi Shrivastava$^{*}$ \quad Piero Kauffmann \quad Ahmed Awadallah \quad Dimitris Papailiopoulos}\\[2pt]
{\normalsize\bf Microsoft Research}\\[6pt]

\end{center}
\begingroup
\renewcommand{\thefootnote}{$^{*}$}
\footnotetext{Correspondence to: \texttt{vaishnavi.shrivastava@microsoft.com}}
\endgroup
\noindent\rule{\textwidth}{0.4pt}

\begin{abstract}
CLI agents are the closest thing language models have to an embodied setting: the model emits commands, the terminal executes them, and the returned stream—stdout, errors, files, logs, and traces—records the consequences. We argue that this stream is a supervision signal, but standard agent RL discards it: GRPO-style training updates action tokens with sparse outcome-level rewards while ignoring environment responses already in the rollout. Failed rollouts provide little policy-gradient signal despite containing rich evidence about how the environment responds. We introduce \textbf{ECHO} (Environment Cross-entropy Hybrid Objective), a hybrid objective that combines the standard policy-gradient loss on action tokens with an auxiliary loss that trains the policy to predict environment observation tokens resulting from its own actions. ECHO reuses the same forward pass as GRPO, requires no additional rollouts, and turns terminal feedback into dense supervision for all rollouts. ECHO doubles GRPO pass@1 on TerminalBench-2.0: Qwen3-8B improves from $2.70\%$ to $5.17\%$, and Qwen3-14B from $5.17\%$ to $10.79\%$. ECHO also produces policies that better predict terminal dynamics, even on trajectories they did not generate: across held-out rollouts, it sharply reduces environment-token cross-entropy while GRPO alone barely changes it. From base Qwen3-8B, ECHO matches expert-SFT-then-GRPO performance on held-out terminal tasks without expert demonstrations, and recovers roughly half of the expert-SFT initialization benefit on TerminalBench-2.0. In some settings, the environment prediction loss alone enables verifier-free self-improvement, allowing policies to improve on unseen OOD tasks by learning only from environment interactions. Together, these results suggest that environment observations are not merely context for future actions, but a dense, on-policy supervision signal already present in every rollout.

\end{abstract}
\vspace{0.6cm}
\begin{figure}[!ht]
  \captionsetup{font=small, labelfont={small,bf}}
  \centering
  \includegraphics[width=0.9\linewidth]{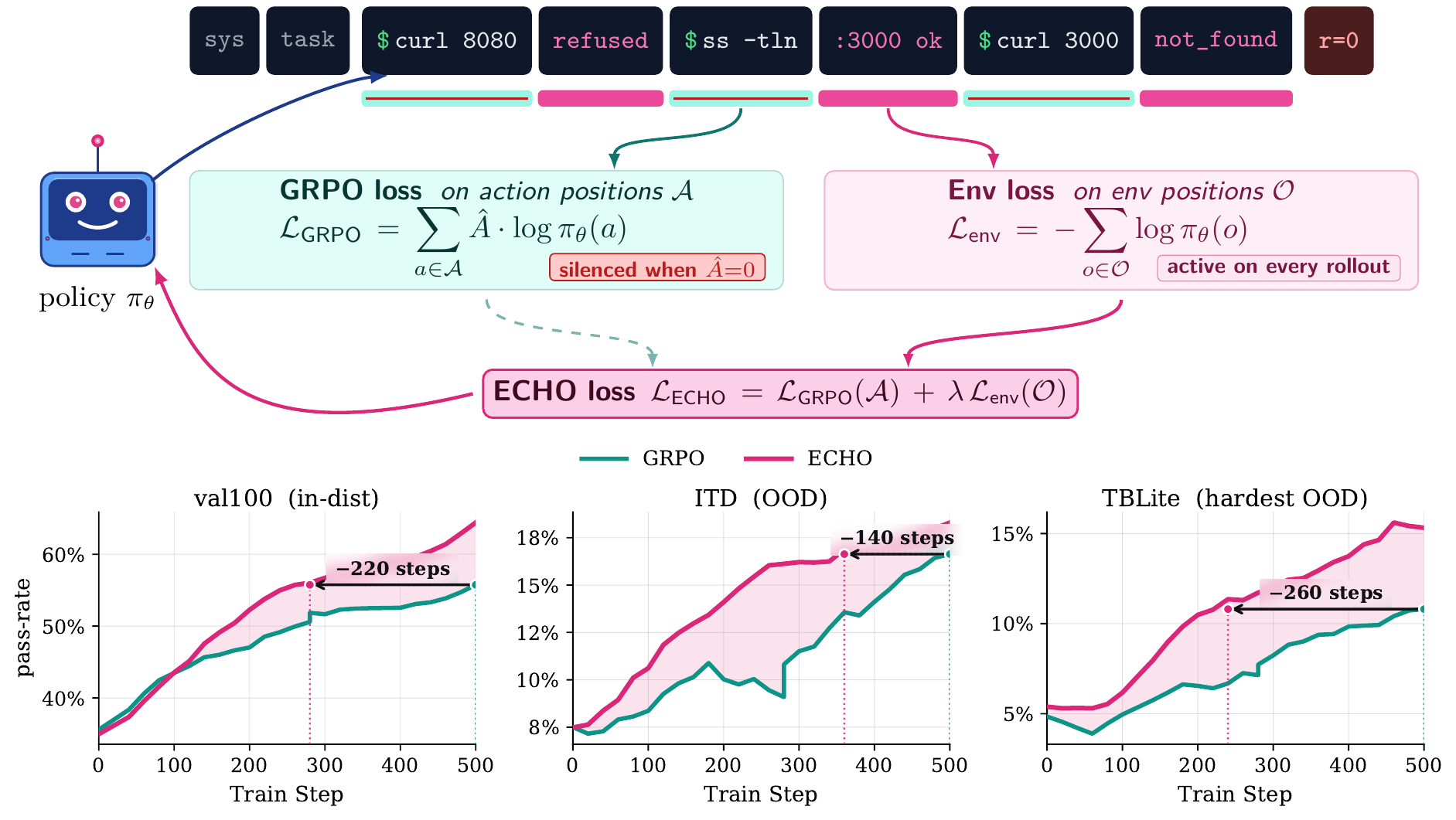}
  \caption{
\textbf{ECHO turns terminal feedback into supervision during agent RL.}
A terminal-agent rollout interleaves assistant actions, with environment observations.
In standard GRPO, previous terminal outputs can inform future commands, but loss is applied only to action-token positions and is driven by sparse outcome-level rewards.
ECHO adds a complementary cross-entropy loss on environment-observation tokens, training the same policy to predict the terminal responses caused by its own actions.
ECHO reuses the same rollout and forward pass while making every terminal response an additional training target.
Bottom: across held-out evaluations, ECHO improves task solve rate relative to matched GRPO-only runs and also reaches the peak GRPO performance substantially earlier in training.
}
  \label{fig:hero}
\end{figure}

\bigskip
\newpage
\section{Introduction}
\label{sec:intro}

Language-model agents learn by acting in environments: a terminal agent edits files, runs tests, reads errors, and issues follow-up commands until a verifier declares success or failure. The terminal responds to every action with real environment feedback, but the reward signal that ultimately reaches the trainer is sparse, delayed, and binary.

This sparsity is particularly pronounced in terminal-agent training. As a concrete example, in our Qwen3-8B setting, often fewer than 15\% of on-policy rollouts solve the task, so under standard GRPO~\citep{Shao2024DeepSeekMathPT} the vast majority of interaction yields little policy-gradient signal. These rollouts are far from uninformative.

Even a failed trajectory contains the actual outputs of whatever the agent ran: file listings, training logs, build errors, the contents of a config file, the response from a web request, a stack trace, the result of a \texttt{grep}, or anything else a shell can produce. Every token from the terminal enters the model's forward computation, yet none of it enters the loss. The transcript becomes context for the next action and nothing more --- and we believe this wastes the most abundant signal in an interactive environment. {\it We instead train on these tokens directly.}

Why should this help? A long-running intuition in language modeling is that good prediction implies good understanding: predicting the next token well, in Sutskever's phrasing, ``means you understand the underlying reality that led to the creation of that token''~\citep{sutskever2023dwarkesh}. We borrow this view for agents. 
More precisely, terminal output is a lossy textual projection of the container state: it reveals stdout, stderr, exit codes, file contents, traces, and test failures, but not the full filesystem, process tree, or hidden task state. Predicting these observations therefore requires the policy to track the latent consequences of its commands: which files were created, which assumptions failed, which tests broke, and what state is likely to be exposed next.
A policy that predicts terminal output well is, in a small but real sense, a policy that understands terminals. 

We introduce \textbf{ECHO} (Environment Cross-entropy Hybrid Objective), a hybrid loss that adds auxiliary cross-entropy on environment-observation tokens to the usual GRPO loss on action tokens for multi-turn agent RL:
\begin{equation}
  \mathcal{L}_{\mathrm{ECHO}}(\theta)
  =
  \mathcal{L}_{\mathrm{GRPO}}(\theta;\mathcal{A})
  +
  \lambda\,\mathcal{L}_{\mathrm{Env}}(\theta;\mathcal{O}'),
  \label{eq:wm}
\end{equation}
where $\mathcal{A}$ indexes assistant-action positions and $\mathcal{O}'$ indexes terminal-output positions inside the environment observations. The objective requires no teacher model, extra rollouts, or an additional forward pass: it uses the same logits already computed for the policy update, but gathers them at a different set of token positions. Because these targets come from the current policy’s own rollouts, ECHO is on-policy: as the agent improves and visits new terminal states, the environment produces new responses to predict, creating a self-evolving curriculum. In effect, ECHO turns the environment stream into dense supervision, so even failed rollouts can teach the policy how the terminal responds.

We test this hypothesis in TerminalBench-style Docker task environments with Qwen3-8B, OpenThinker-Agent-v1-SFT, and Qwen3-14B starting policies. ECHO consistently improves both internal held-out evaluations and the public TerminalBench-2.0 benchmark. On TerminalBench-2.0, ECHO nearly doubles GRPO's pass@1 rate from $2.70\%$ to $5.17\%$ for Qwen3-8B and from $5.17\%$ to $10.79\%$ for Qwen3-14B. The same checkpoints also become substantially better predictors of terminal behavior: on held-out off-policy trajectories from Qwen3-32B, ECHO sharply lowers environment-token cross-entropy while GRPO alone barely changes it. ECHO also reduces dependence on expert demonstrations: from a base Qwen3-8B, ECHO matches OpenThinker-SFT+GRPO on internal evaluations without using any of the ${\sim}15$k expert demonstrations behind the SFT model, and closes about half of the expert-SFT gap on TerminalBench-2.0.

Our contributions are as follows:
\begin{enumerate}[nosep, leftmargin=*, labelindent=0pt, labelsep=0.4em]
  \item \textbf{ECHO turns terminal outputs into supervision.} We introduce an on-policy hybrid objective that treats terminal outputs — stdout, errors, files, and tool traces — as dense training targets, adding environment-token cross-entropy to GRPO’s action-token loss. The two terms share one forward pass, require no teacher model, and add no extra rollouts. (\S\ref{sec:method})
  \item \textbf{Consistent improvements over GRPO.} ECHO improves Qwen3-8B, OpenThinker-Agent-v1-SFT, and Qwen3-14B on both internal evaluations and TerminalBench-2.0, nearly doubling pass@1 at 8B and 14B. (\S\ref{sec:headline})
  \item \textbf{ECHO learns terminal dynamics.} On held-out trajectories from a stronger Qwen3-32B policy, ECHO sharply lowers environment-token cross-entropy across model families and evaluation slices, indicating better prediction of terminal behavior, while GRPO alone barely changes it. (\S\ref{sec:mechanism})
  \item \textbf{Reduced reliance on expert SFT.} From a base Qwen3-8B, ECHO matches SFT-bootstrapped GRPO, without using any expert demonstrations. (\S\ref{sec:expert-sft})
\item \textbf{Verifier-free adaptation.} Even without a verifier, a policy can sometimes improve just by interacting with the environment and predicting the environment's response to its own actions. (\S\ref{sec:wm-only})
\end{enumerate}
Taken together, these findings suggest agent training has been operating with a supervision source masked out: every policy action has a consequence in the environment, and that consequence is in the rollout already. ECHO shows that these consequences can be trained on directly, turning even failed interactions into signal for learning how the world responds.
\section{Preliminaries}
\label{sec:prelim}

\paragraph{Multi-Turn Rollout Structure.}
A training sequence interleaves a system prompt, the user task, and a transcript of (assistant action, environment observation) pairs:
\[
\underbrace{[\text{sys}]\,[\text{task}]}_{\text{prompt}}\;
[\text{action}_1]\,[\text{obs}_1]\,
[\text{action}_2]\,[\text{obs}_2]\,
\cdots\,[\text{action}_K]\,[\text{obs}_K].
\]
At each turn, the policy samples action tokens conditioned on the entire prior transcript; the harness parses these into a bash command, executes it in the container, and appends the resulting terminal output as the next observation. Let $\mathcal{A}\subseteq\{1,\dots,T\}$ index assistant-action positions and $\mathcal{O}\subseteq\{1,\dots,T\}$ index environment-observation positions. Trainers compute log-probabilities on the full sequence (every action depends on prior observations), but the policy-gradient loss is applied only on $\mathcal{A}$. Observations in $\mathcal{O}$ are conditioned on, but receive no direct training signal.

\paragraph{Group-Relative Policy Optimization.}
GRPO~\citep{Shao2024DeepSeekMathPT} optimizes a clipped policy-gradient objective with group-normalized advantages and no learned value function. For prompt $x$, sampled rollouts $\{y^{(i)}\}_{i=1}^n$, and binary rewards $r^{(i)}\in\{0,1\}$, each rollout receives a scalar group-normalized advantage $\hat A^{(i)}$, applied uniformly to its action-token positions $\mathcal{A}$ using the clipped importance ratio $\rho_t^{(i)}$:
\begin{equation}
  \mathcal{L}_{\mathrm{GRPO}}(\theta; \mathcal{A})
  =
  -\frac{1}{\sum_i |\mathcal{A}^{(i)}|}
  \sum_i \sum_{t\in\mathcal{A}^{(i)}}
  \min\!\left(
    \rho_t^{(i)}\hat A^{(i)},
    \mathrm{clip}(\rho_t^{(i)},1-\epsilon,1+\epsilon)\hat A^{(i)}
  \right).
  \label{eq:grpo}
\end{equation}
GRPO optimizes only assistant action tokens. Observation tokens remain in context and affect future actions, but are not policy-gradient targets. With sparse binary rewards, all-zero groups have no reward contrast. In mixed groups, unsuccessful trajectories receive only a trajectory-level negative signal, so learning concentrates on rare successful rollouts.

\section{Method}
\label{sec:method}

\subsection{ECHO Objective}

ECHO is a hybrid loss objective combining GRPO's policy gradient loss on action tokens with a supervised next-token objective on observation tokens. Let $p_\theta(\cdot\mid x_{<t})$ denote the model's next-token distribution. The ECHO loss augments GRPO with an Environment-Prediction Loss: length-normalized cross-entropy on a subset $\mathcal{O}'\subseteq\mathcal{O}$ of observation tokens: 
\begin{equation}
  \mathcal{L}_{\text{Env}}(\theta;\,\mathcal{O}')
  \;=\; -\,\frac{1}{Z}\sum_{t\in\mathcal{O}'} \log p_\theta(x_t\mid x_{<t}),
  \label{eq:env-loss}
\end{equation}
where $Z = |\mathcal{O}|$ normalizes each sequence by its total observation length. We normalize by the total observation length $|\mathcal{O}|$, rather than $|\mathcal{O}'|$, so runs with different target subsets remain comparable on a per-observation scale.
ECHO is the joint objective $\mathcal{L}_{\text{total}}=\mathcal{L}_{\text{GRPO}}+\lambda\mathcal{L}_{\text{Env}}$ as in equation~\ref{eq:wm}.  Because the observation targets come from the current policy's own rollouts, ECHO is on-policy: as the agent improves and visits new terminal states, the environment-prediction targets evolve with the policy rather than remaining a frozen offline set of trajectories.

The two losses share a single actor forward pass: the same logits feed both, gathered through different masks (assistant-action positions for GRPO, additional observation positions for ECHO's Environment-Prediction loss). ECHO therefore requires no second rollout, teacher model, or second forward pass; the only added work is a masked log-probability sum. The intended effect is representation shaping: by learning which observations follow from which commands, the same policy network can develop better priors over which future commands are likely to expose useful state, repair errors, or advance the task. It composes orthogonally with stabilization techniques aimed at the policy gradient itself, including void-trajectory filtering~\citep{xue2025simpletir}, overlong filtering and clip-higher~\citep{Yu2025DAPOAO}, and KL-regularized reference updates~\citep{rlhf-instruct2022}. Algorithm~\ref{alg:wm} summarizes the resulting update.

\begin{algorithm}[ht]
\caption{ECHO}
\label{alg:wm}
\setlength{\fboxsep}{5pt}
\noindent\fcolorbox{echoline}{echobg}{%
  \begin{minipage}{0.96\linewidth}
  \begin{algorithmic}[1]
  \Require rollout sequence $x_{1:T}$; action mask $\mathcal{A}$; env-observation mask $\mathcal{O}'$; full-observation length $|\mathcal{O}|$; advantages $\{\hat A_t\}_{t\in\mathcal{A}}$; mixing coefficient $\lambda$.
  \State $\boldsymbol{\ell} \gets \texttt{forward}_\theta(x_{1:T})$ \Comment{single actor forward; logits at every position}
  \State $\log p_t \gets \log\,\text{softmax}(\boldsymbol{\ell}_t)[x_t]$ for all $t$
  \State $\mathcal{L}_{\text{GRPO}} \gets \texttt{ClippedGRPO}\bigl(\{\log p_t\}_{t\in\mathcal{A}},\,\{\hat A_t\}_{t\in\mathcal{A}}\bigr)$
  \State $\mathcal{L}_{\text{Env}} \gets -\,\bigl(\textstyle\sum_{t\in\mathcal{O}'} \log p_t\bigr) / |\mathcal{O}|$
  \State \Return $\mathcal{L}_{\text{GRPO}} + \lambda\,\mathcal{L}_{\text{Env}}$
  \end{algorithmic}
  \end{minipage}%
}
\end{algorithm}

Computationally, ECHO changes the loss mask rather than the rollout or model evaluation. The expensive attention and MLP computations already run over the full rollout to compute action-token log-probabilities for GRPO. ECHO simply gathers the already-computed logits at terminal-output positions and includes their cross-entropy in the same backward pass.

\subsection{Choosing Observation Targets}
\label{sec:method-target}

We set $\mathcal{O}'$ to the env tokens only, excluding the harness's warning prefix. An observation message has internal structure
\[
\underbrace{\text{\texttt{WARNINGS:\textbackslash n- ...}}}_{\text{warning tokens}}\;\;
\underbrace{\text{\texttt{<command\_output> ... </command\_output>}}}_{\text{terminal-output (env) tokens}},
\]
where the warning block is a rule-based message emitted when the previous tool call fails parsing or violates format constraints, and the env block carries the actual command output. The reason for excluding warnings is empirical: warning tokens are low-entropy and the model memorizes them within $\sim$60 training steps, so warn-only configurations quickly lose useful gradient. Terminal-output tokens, by contrast, encode task-specific feedback (file names, test failures, byte counts, error formats) and continue to provide informative gradient throughout training.

\subsection{Tuning the Loss Weight}
We swept $\lambda\in\{0.001,0.005,0.01,0.02,0.05,0.1,0.2\}$ and found a productive range of $0.01$--$0.05$. Below this range, the auxiliary gradient is too small to shape representations reliably: environment prediction loss can fluctuate or increase while the policy objective dominates. Above this range, the observation objective begins to compete with the policy update; at $\lambda=0.1$ policy quality plateaus or degrades, and at $\lambda=0.2$ runs can collapse into degenerate rollouts whose terminal outputs are easy to predict but no longer useful. We therefore use a constant $\lambda=0.05$ in all reported experiments. The constant weight is naturally self-annealing: as the model learns terminal-output statistics, $\mathcal{L}_{\text{Env}}$ falls rapidly, reducing the auxiliary contribution without an explicit schedule.

\section{Experimental Setup}
\label{sec:setup}
\paragraph{Training Task Corpus.}
We start from 2700 curated terminal tasks: 1977 from Endless Terminals~\citep{endlessterminals2026,endlessterminalsdata2026} and 723 from OpenThoughts-Agent-v1-RL~\citep{openthoughtsrl2025}, after filtering out analysis/computation, specialized-application, infrastructure/networking, and complex-bash domains. We then generate 6170 additional tasks with a modified Endless Terminals pipeline, covering task specification, Dockerfile generation/validation, and Harbor-format export. We retain only tasks solved by GPT-5 in at least one of 16 attempts, yielding 8870 tasks across data processing, system operations, and development/tooling. We train on 8770 tasks and hold out 100 for in-distribution validation (\textbf{val100}).

\paragraph{Harness and Runtime Environment.}
At each turn, the policy conditions on its prior reasoning, commands, and command outputs, then emits a thinking block followed by Qwen XML-format bash commands or a task-done signal. A minimal training harness parses the first command or completion signal, executes it, and returns optional format warnings plus stdout/stderr and exit code as the next observation. Episodes run for up to 16 turns in Docker, orchestrated by Harbor~\citep{harbor2025}, with a 16k context window and at most 2048 generated tokens per turn. We verify success with unit tests at episode end, using 10-minute agent and 2-minute verifier timeouts per training task.

\paragraph{Models.}
We train on three starting policies: \textbf{Qwen3-8B}~\citep{qwen3-2025},
\textbf{OpenThinker-Agent-v1-SFT} (OT-SFT)~\citep{openthinker2025}, and
\textbf{Qwen3-14B}~\citep{qwen3-2025}. OT-SFT is a Qwen3-8B model SFT'd on
${\sim}15\text{k}$ expert terminal-agent demonstrations from the GLM-4.6 model.

\paragraph{RL Recipe.}
All experiments use the same GRPO recipe: $n=16$ rollouts per prompt, batch size of 16, learning rate $1\times10^{-6}$, gradient clip $0.2$, prompt-level advantage normalization, sequence-level loss aggregation, no KL penalty unless noted, and rollout temperature $0.8$. For ECHO runs, we use $\lambda=0.05$ to scale the Environment-Prediction loss. We provide a reward of 1 if final tests for a task pass, and a reward of 0 otherwise. We train each model for 500 GRPO steps on 8 B200 GPUs.

\paragraph{Evaluation.}
We evaluate model performance on \textbf{val100}, internal-dev (\textbf{ITD}), OpenThoughts-TBLite 
(\textbf{TBLite})\linebreak[1]~\citep{openthoughtstblite2025}, and \textbf{TerminalBench-2.0} (TB2)~\citep{terminalbench2025}. 
val100 is a held-out set of 100 tasks from our training corpus. Internal-dev is a set of 71 tasks focusing on data processing, systems operations, and development/tooling, sampled from TerminalBench 1.0 (core and non-core) and OpenThoughts-TB-dev~\citep{openthoughtstbdev2025}. OpenThoughts-TBLite is a set of 100 terminal-bench-style tasks calibrated for small-model performance relative to the harder TB2 benchmark. On val100, ITD, and TBLite, we evaluate using our minimal agent harness, with 8 rollouts per task at temperature 0.6. On TB2, we use the Terminus 2 harness~\citep{terminus2025} and perform 5 rollouts at temperature 0.6 with 32k context.

\section{Results}
\label{sec:results}
\begin{table}[!t]
\centering
\small
\setlength{\tabcolsep}{4.5pt}
\begin{tabular}{ll@{\hskip 8pt}cc!{\color{black!40}\vrule width 0.6pt}cccc}
\toprule
\textbf{Model} & \textbf{Setup} & \textbf{val100} & \textbf{ITD} & \textbf{TBLite} & \textbf{TB2 p@1} & \textbf{TB2 p@3} & \textbf{TB2 p@5} \\
\midrule
\rowcolor{rowbase} Qwen3-8B  & Base                  & 34.2 &  7.0 &  4.9 &  1.57 &  3.71 &  4.49 \\
\rowcolor{rowrl}             & GRPO             & 54.9 & 16.2 &  9.5 &  2.70 &  6.74 &  8.99 \\
\rowcolor{rowecho}           & ECHO  & \textbf{63.7} & \textbf{18.9} & \textbf{11.4} & \textbf{5.17} & \textbf{10.45} & \textbf{13.48} \\
\midrule
\rowcolor{rowbase} OT-SFT    & SFT                   & 38.5 & 10.7 &  6.0 &  5.62 & 10.45 & 12.36 \\
\rowcolor{rowrl}             & GRPO             & 63.5 & 18.8 & 11.6 &  7.64 & \textbf{14.38} & 17.98 \\
\rowcolor{rowecho}           & ECHO  & \textbf{73.1} & \textbf{22.7} & \textbf{12.9} & \textbf{7.87} & 13.82 & \textbf{17.98} \\
\midrule
\rowcolor{rowbase} Qwen3-14B & Base                  & 35.3 & 12.1 &  5.7 &  4.27 &  8.99 & 12.36 \\
\rowcolor{rowrl}             & GRPO             & 60.3 & 17.9 &  9.8 &  5.17 & 10.67 & 13.48 \\
\rowcolor{rowecho}           & ECHO  & \textbf{65.0} & \textbf{19.8} & \textbf{15.1} & \textbf{10.79} & \textbf{16.52} & \textbf{19.10} \\
\bottomrule
\end{tabular}
\vspace{5pt}
\captionsetup{font=small, labelfont={small,bf}}
\caption{\textbf{Pass rates before RL, after GRPO, and after ECHO for three starting policies.} For \emph{val100} (100 tasks), \emph{ITD} (71 tasks), \emph{TBLite} (OpenThoughts-TBLite, 100 tasks) we report pass@1 over 8 attempts. For \emph{TB2} (TerminalBench-2.0, 89 tasks) we report pass@$k$ over 5 attempts. Bold marks the best metrics in each block.}
\label{tab:headline}
\end{table}

\providecommand{\resultfinding}[1]{%
  \par\smallskip
  \setlength{\fboxsep}{6pt}%
  \noindent\fcolorbox{echoline}{echobg}{%
    \begin{minipage}{\dimexpr\linewidth-2\fboxsep-2\fboxrule\relax}
    \small #1
    \end{minipage}%
  }%
  \par\smallskip
}

ECHO improves every starting policy on every benchmark we test.
TerminalBench-2.0 pass@1 nearly doubles at both 8B ($2.70\to5.17$) and 14B
($5.17\to10.79$), and internal pass rates rise on every slice. The same
checkpoints become substantially better predictors of terminal feedback.
They match GRPO's peak performance in $1.5$--$2.3\times$ fewer training steps and
waste fewer turns at inference. Starting from base Qwen3-8B, ECHO fully matches what an expert SFT initialization buys on internal
evaluations, and recovers half of its lead on TerminalBench-2.0 --- without
using any of the ${\sim}15$k expert demonstrations the SFT model requires.

\subsection{ECHO Improves Over GRPO Performance}
\label{sec:headline}

\resultfinding{\textbf{ECHO consistently improves task success.}
ECHO raises task success on every internal evaluation (val100 and ITD), TBLite, and TerminalBench-2.0.
TerminalBench-2.0 pass@1 nearly doubles
for Qwen3-8B ($2.70\to5.17$, $\times1.9$) and Qwen3-14B ($5.17\to10.79$,
$\times2.1$).}

Table~\ref{tab:headline} compares matched GRPO and ECHO checkpoints across three starting policies.
The setup isolates a single change: whether terminal-output tokens are additionally trained with a
cross-entropy objective alongside the standard policy-gradient loss. Across all three starting
policies, ECHO improves every internal evaluation metric and consistently boosts performance on
TerminalBench-2.0 under the Terminus-2 harness. At 8B, TB2 pass@1 nearly doubles from
2.70 to 5.17; at 14B, it rises from 5.17 to 10.79, with pass@5 increasing from 13.48 to 19.10.

\begin{figure}[t]
  \centering
  \includegraphics[width=\linewidth]{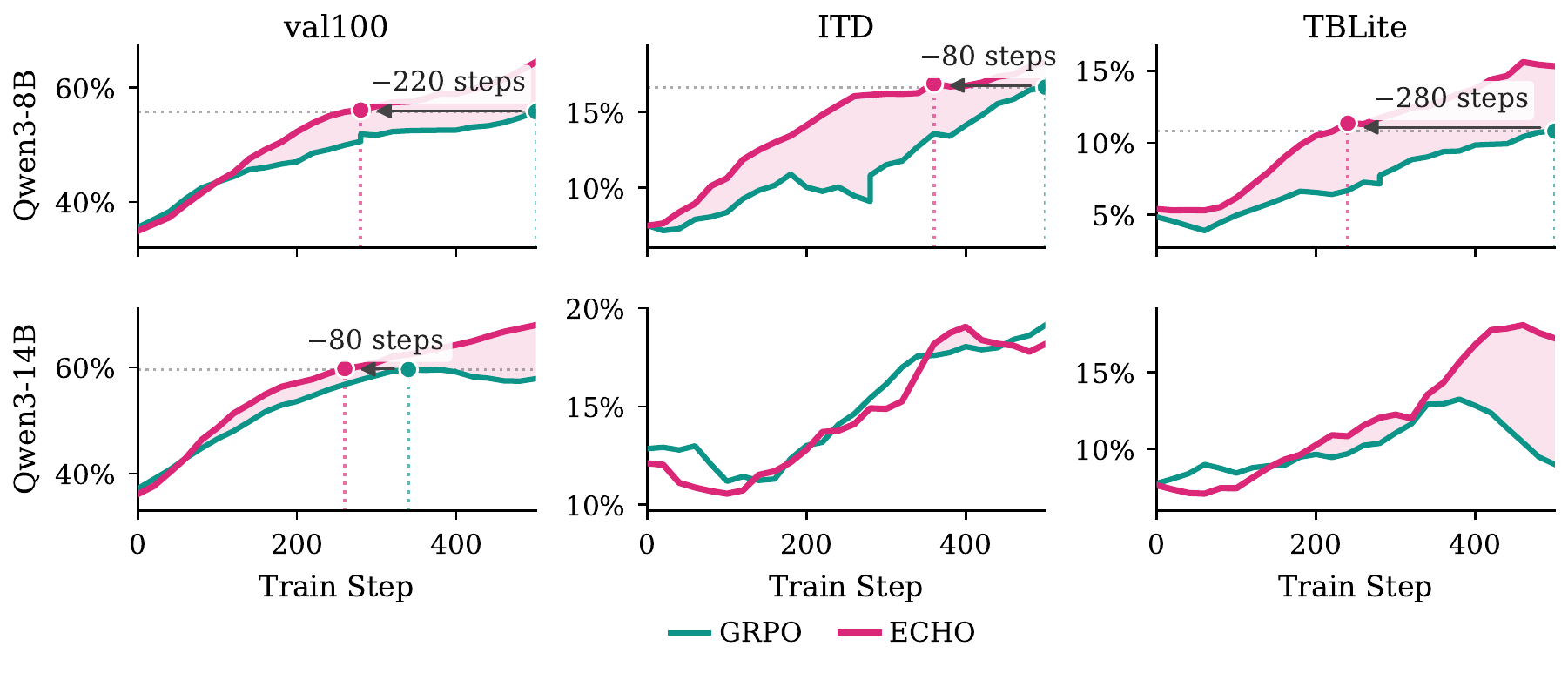}
  \captionsetup{font=small, labelfont={small,bf}}
  \caption{\textbf{Pass-rate training curves over 500 GRPO steps.} Teal shows GRPO only; pink shows ECHO, with shading where ECHO exceeds the matched GRPO baseline. Rows are Qwen3-8B base and Qwen3-14B base; columns are val100 (in-distribution), ITD (OOD), and TBLite (hardest OOD). OT-SFT--initialized curves appear in App.~\ref{app:otsft-curves}.}
  \label{fig:curves-q8b-q14b}
\end{figure}

The 14B result is particularly notable. Although the internal gains at 14B are smaller than at 8B,
the improvements on TerminalBench-2.0 are substantially larger. One plausible explanation is
that the larger model can internalize more generalizable terminal dynamics from the observation
stream, while at smaller scales the policy and environment-prediction objectives compete more directly
for limited capacity. Figure~\ref{fig:curves-q8b-q14b} shows the corresponding training dynamics: at 8B, ECHO
consistently outperforms the GRPO baseline throughout training, while at 14B it reaches
a substantially higher final plateau.

\subsection{Does ECHO Really Learn Terminal Dynamics?}
\label{sec:mechanism}
A useful terminal dynamics model should be predictive: given an action, it should be able to simulate the environment’s response. We test this directly by measuring environment-token cross-entropy on held-out trajectories: the likelihood the policy assigns to the terminal-output tokens that actually follow each action. We measure how well each model predicts terminal-output tokens on off-policy trajectories generated by a stronger Qwen3-32B model. Across val100, ITD, and
TBLite, we evaluate on 8 trajectories per task, totaling 2,168 trajectories.

This evaluation is intentionally off-policy: the evaluated models did not generate these
trajectories themselves. Low cross-entropy therefore requires predicting the outcomes of
another stronger agent's actions, rather than memorizing a model's own rollout distribution. In this sense, environment-token cross-entropy provides an operational test of the world-modeling claim: a model that has learned more about terminal dynamics should better simulate terminal responses, even on trajectories it did not generate.

\begin{figure}[t]
  \centering
  \includegraphics[width=\linewidth]{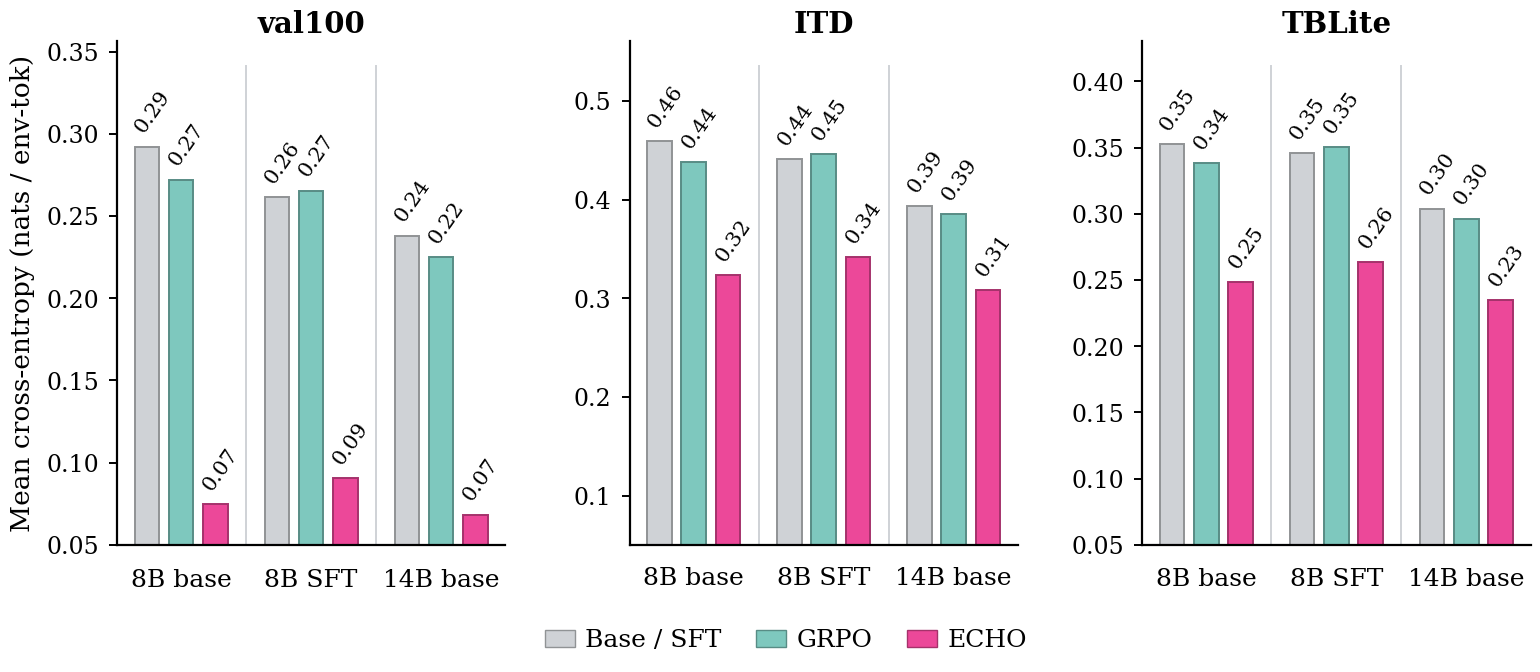}
  \captionsetup{font=small, labelfont={small,bf}}
  \caption{\textbf{Per-token cross-entropy on terminal-output tokens} for trajectories from a stronger model, Qwen3-32B. Each panel evaluates a different distribution; within each panel, starting policies are shown before RL, after GRPO, and after ECHO. GRPO barely changes env-token CE relative to the starting policy, whereas ECHO sharply lowers it across all model families and evaluation slices. Lower is better.}
  \label{fig:env-ce}
\end{figure}

\resultfinding{\textbf{ECHO learns transferable terminal dynamics.}
On held-out, off-policy trajectories from Qwen3-32B, ECHO sharply lowers environment-token
cross-entropy across all starting policies and evaluation slices, while GRPO alone
barely changes it.
}

Figure~\ref{fig:env-ce} shows exactly this pattern. GRPO alone barely changes
environment-token cross-entropy relative to the starting policy, despite improving task success.
ECHO, by contrast, sharply lowers prediction error across all starting policies and evaluation
distributions. For Qwen3-14B, cross-entropy drops from 0.24$\to$0.07 on val100,
0.39$\to$0.31 on ITD, and 0.30$\to$0.23 on TBLite; for Qwen3-8B, the corresponding drops
are 0.29$\to$0.07, 0.46$\to$0.32, and 0.35$\to$0.25. The larger reduction on val100 is expected: val100 is drawn from the same task distribution as training, whereas ITD and TBLite are out-of-distribution evaluations, so successful transfer requires predicting terminal behavior under less familiar task structure.
These results support the central mechanism behind ECHO: the environment-prediction objective
improves the policy's ability to simulate terminal responses, and this ability transfers beyond the
model's own trajectories.

\subsection{ECHO Reduces Dependence on Expert Demonstrations}
\label{sec:expert-sft}
Expert SFT primes terminal agents before RL by behavior-cloning demonstrations from a stronger policy. In our comparison, OT-SFT is Qwen3-8B SFT'd on $\sim$15k expert demonstrations from
a GLM-4.6 teacher. We ask how much of this expert initialization can be replaced by letting the
base model explore and learn from its own terminal interactions.
\begin{wrapfigure}{r}{0.46\linewidth}
    \centering
    \includegraphics[width=0.8\linewidth]{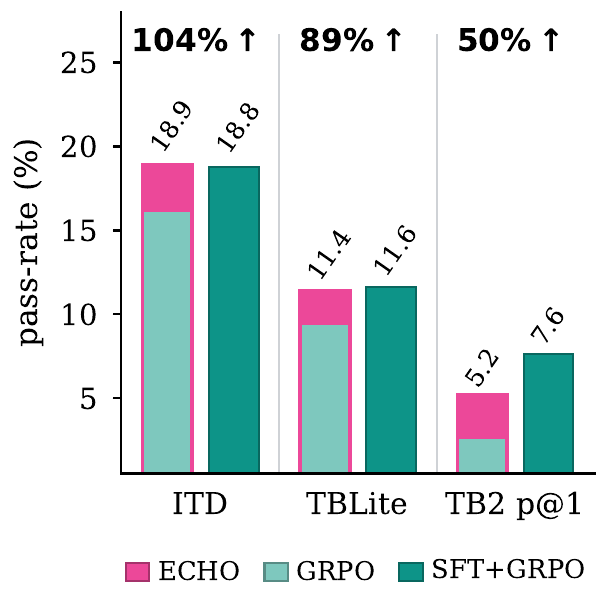}
    \captionsetup{font=small, labelfont={small,bf}}
    \caption{\textbf{ECHO recovers most of the benefit of expert-SFT} initialization on ITD and TBLite and roughly half on TerminalBench-2.0.}
    \label{fig:sft-gap-narrow}
    \vspace{-2 em}
  \end{wrapfigure}
We define the expert-SFT gap as the gain from OT-SFT+GRPO over Qwen3-8B+GRPO, and
the ECHO lift as the gain from ECHO over Qwen3-8B+GRPO. \label{sec:sftgap}
 Figure~\ref{fig:sft-gap-narrow}
shows that ECHO recovers almost all of the OT-SFT advantage on internal evaluations:
101.6\% of the gap on val100, 103.9\% on ITD, and 88.9\% on TBLite. This suggests that much of what expert SFT provides is an interaction prior: how terminal agents inspect files, run tests, encounter errors, follow tracebacks, and expose useful state. ECHO does not imitate the expert’s action choices; instead, it learns from the terminal consequences of the base model’s own actions. This can recover the interaction-modeling component of expert SFT without behavior-cloning the teacher.

On TerminalBench-2.0, ECHO closes roughly half of the OT-SFT gap: 50.0\% on pass@1,
48.6\% on pass@3, and 50.0\% on pass@5. This is weaker than the internal recovery but still
substantial. A likely explanation is that TB2 requires not only terminal familiarity, but also
higher-level strategy. ECHO narrows the gap by learning the interaction prior directly from environment feedback, but expert demonstrations still help with the strategy prior: which commands to try first, how to decompose a task, when to inspect versus edit, and when to stop. Thus, ECHO does not make expert demonstrations obsolete. Rather, it suggests that one component of their value—familiarity with terminal feedback and state evolution—can be learned directly from interaction.
\resultfinding{\textbf{ECHO substantially closes the expert-SFT gap.}
On Qwen3-8B, ECHO recovers nearly all of the OT-SFT initialization gain on internal evaluations and about half of the TerminalBench-2.0 gain, without behavior-cloning expert trajectories.
}

\subsection{Training and Inference Efficiency}
\label{sec:efficiency}

Because ECHO trains on observation tokens already present in each rollout, it can improve how much
learning the policy extracts from a fixed sample budget. We measure this by asking when each ECHO
run first exceeds the best validation score reached by its matched GRPO-only run within 500 steps, using
the aggregate total score (val100+ITD+TBLite) and TBLite alone.

\resultfinding{\textbf{ECHO often learns faster and uses inference budget more productively.}
At 8B, ECHO reaches the GRPO-only peak in $1.5$--$2.3\times$ fewer training steps. At 14B,
both runs peak at the same step, but ECHO reaches a higher plateau. At inference, ECHO halves
TB2 timeouts for Qwen3-8B and OT-SFT and reduces completion tokens for all three model pairs.
}

\begin{table}[t]
\centering
\small
\setlength{\tabcolsep}{4.5pt}
\begin{tabular}{l
  >{\columncolor{rowrl}}c >{\columncolor{rowecho}}c c
  >{\columncolor{rowrl}}c >{\columncolor{rowecho}}c c}
\toprule
& \multicolumn{3}{c}{\textbf{Total}}
& \multicolumn{3}{c}{\textbf{TBLite}} \\
\cmidrule(lr){2-4}\cmidrule(lr){5-7}
\textbf{Model} & GRPO & ECHO & Speedup & GRPO & ECHO & Speedup \\
\midrule
Qwen3-8B   & 460 & 240 & \cellcolor{rowecho}$1.92\times$ & 500 & 220 & \cellcolor{rowecho}$2.27\times$ \\
OT-SFT     & 400 & 260 & \cellcolor{rowecho}$1.54\times$ & 220 & 300 & \cellcolor{rowrl}$0.73\times$ \\
Qwen3-14B  & 380 & 380 & $1.00\times$                    & 380 & 380 & $1.00\times$ \\
\bottomrule
\end{tabular}
\vspace{5pt}
\captionsetup{font=small, labelfont={small,bf}}
\caption{\textbf{First step at which each run reaches its best internal score within 500 GRPO steps.} \emph{Total} aggregates val100, ITD, and TBLite; \emph{TBLite} isolates the hardest OOD slice. Pink speedup cells indicate ECHO reaches the score faster ($>1\times$); teal indicates GRPO is faster.}
\label{tab:efficiency}
\end{table}

Table~\ref{tab:efficiency} shows large 8B speedups on the aggregate score
($1.54$--$1.92\times$) and for Qwen3-8B on TBLite ($2.27\times$). The exception is
OT-SFT on TBLite, where GRPO peaks earlier. At 14B, ECHO and GRPO peak at the same step, but
ECHO is higher throughout training (Fig.~\ref{fig:curves-q8b-q14b}, bottom row), so the benefit
appears as a higher plateau rather than faster convergence.

\begin{table}[t]
\centering
\small
\setlength{\tabcolsep}{3.5pt}
\begin{tabular}{l
  >{\columncolor{rowrl}}c >{\columncolor{rowecho}}c c
  >{\columncolor{rowrl}}c >{\columncolor{rowecho}}c c
  >{\columncolor{rowrl}}c >{\columncolor{rowecho}}c c}
\toprule
& \multicolumn{3}{c}{\textbf{Timeout rate (\%)}}
& \multicolumn{3}{c}{\textbf{Turns}}
& \multicolumn{3}{c}{\textbf{Tokens}} \\
\cmidrule(lr){2-4}\cmidrule(lr){5-7}\cmidrule(lr){8-10}
\textbf{Model} & GRPO & ECHO & $\Delta$ & GRPO & ECHO & $\Delta$ & GRPO & ECHO & $\Delta$ \\
\midrule
Qwen3-8B   & 19.8 & 9.0  & \cellcolor{rowecho}$-55\%$ & 24.3 & 19.8 & \cellcolor{rowecho}$-18\%$ & 43.2k & 30.3k & \cellcolor{rowecho}$-30\%$ \\
OT-SFT     & 45.2 & 24.7 & \cellcolor{rowecho}$-45\%$ & 66.3 & 37.7 & \cellcolor{rowecho}$-43\%$ & 35.4k & 34.8k & \cellcolor{rowecho}$-2\%$ \\
Qwen3-14B  & 40.7 & 43.1 & \cellcolor{rowrl}$+6\%$    & 22.0 & 23.5 & \cellcolor{rowrl}$+7\%$    & 49.8k & 43.1k & \cellcolor{rowecho}$-13\%$ \\
\bottomrule
\end{tabular}
\vspace{5pt}
\captionsetup{font=small, labelfont={small,bf}}
\caption{\textbf{TerminalBench-2.0 inference-time trajectory statistics.} \emph{Timeout rate}: fraction of TB2 trials that hit the per-task time limit of 1200 seconds. \emph{Turns}: average agent turns per trial. \emph{Tokens}: average completion tokens per trial. Pink-shaded $\Delta$ cells indicate ECHO wins (lower is better on all three columns); teal indicates GRPO wins.}
\label{tab:inference-efficiency}
\end{table}

Table~\ref{tab:inference-efficiency} shows that ECHO also improves how agents spend their
inference budget. For Qwen3-8B, ECHO cuts TB2 timeouts from $19.8\%$ to $9.0\%$ and
completion tokens by $30\%$; for OT-SFT, it cuts timeouts from $45.2\%$ to $24.7\%$ and turns
by $43\%$. Qwen3-14B is the only exception on timeouts and turns, but still uses $13\%$ fewer
tokens while achieving the largest TB2 pass@1 gain. Overall, ECHO produces policies that
not only solve more tasks, but spend less interaction doing so.

\subsection{Verifier-Free Adaptation}
\label{sec:wm-only}

The main results use ECHO as a joint objective that adds environment-observation prediction during RL. We next ask whether
the observation loss alone can improve a policy on unseen tasks, without unit-test
rewards or any policy-gradient signal. Starting from our strongest 8B ECHO
checkpoint, we mask out the GRPO term and continue training for 100 steps using only 
$\mathcal{L}_{\mathrm{Env}}$ on environment tokens. The model still acts in the environment, observes the terminal response, and updates only by predicting the terminal-output tokens caused by its own actions. Since there is no label indicating if a trajectory was correct or an action was good, any task improvement must arise indirectly: predicting terminal feedback must reshape the model states from which future actions are sampled.

\resultfinding{\textbf{Environment prediction can improve agents without verifier.}
On unseen in-distribution tasks, env-only adaptation improves val100 by $+3.8$pp. On harder OOD tasks, after filtering to clean tool-call trajectories, it improves PyTerm by $+10.0$pp and ITD by $+5.2$pp without any reward signal, while preserving
val100 within $\pm1$pp.}

We evaluate verifier-free adaptation on val100, ITD, TBLite, and PyTerm, a held-out
set of 928 synthetic terminal tasks emphasizing Python script generation. For PyTerm,
we use 828 tasks for env-only adaptation and 100 for evaluation. The starting
checkpoint solves $\sim$11.3\% of PyTerm, comparable to its TBLite pass rate. In all cases, adaptation uses no task rewards, verifier outputs, or action-token loss.

\begin{table}[t]
\centering
\small
\setlength{\tabcolsep}{6pt}
\begin{tabular}{llcrr}
\toprule
\textbf{Target dist.} & \textbf{Rollout filter} & \textbf{Step} & \textbf{$\Delta$ target (pp)} & \textbf{$\Delta$ val100 (pp)} \\
\midrule
val100 (in-dist.) & none           & 70  & $+3.8$  & $+3.8$ \\
PyTerm (OOD)      & clean tool calls & 100 & $+10.0$ & $-0.9$ \\
ITD (OOD)         & clean tool calls & 100 & $+5.2$  & $+0.4$ \\
TBLite (OOD)      & clean tool calls & 100 & $-3.9$  & $-0.4$ \\
\bottomrule
\end{tabular}
\vspace{5pt}
\label{tab:wm-only}
\end{table}

\begin{figure}[t]
  \centering
  \includegraphics[width=\linewidth]{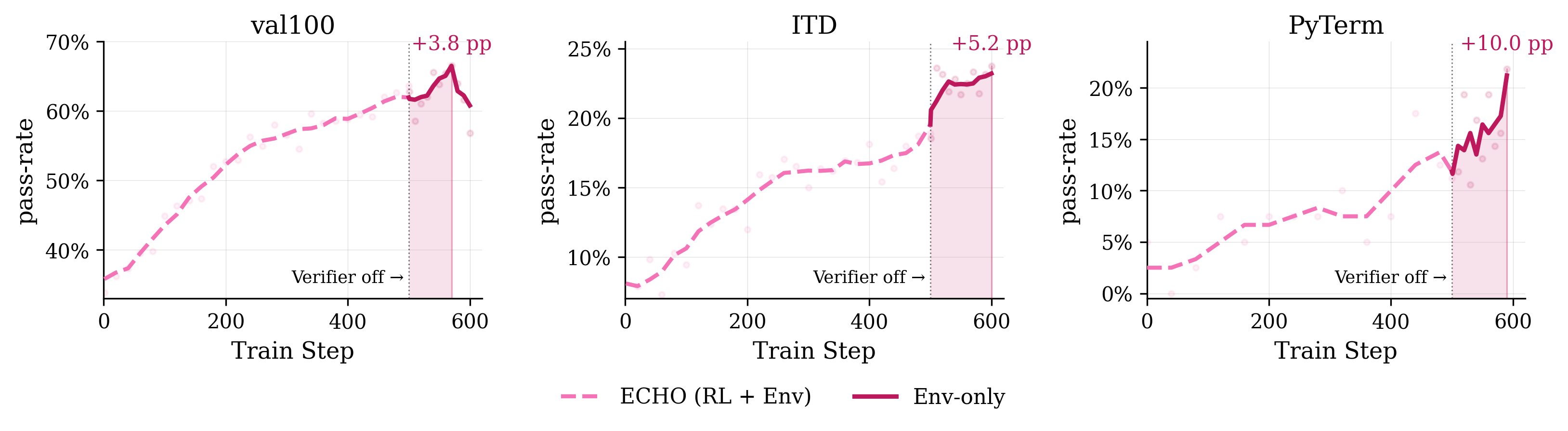}
  \captionsetup{font=small, labelfont={small,bf}}
  \caption{\textbf{Verifier-free adaptation from environment prediction alone.} Starting from the strongest Qwen3-8B ECHO checkpoint, we turn the verifier off at step 500 and continue training only on environment-token cross-entropy. The table reports the best post-adaptation change for each target distribution: ``clean tool calls'' means retaining only rollouts where every tool call is parseable and valid. Env-only adaptation improves val100 without filtering (+3.8 pp), and improves ITD (+5.2 pp) and PyTerm (+10.0 pp) after clean-rollout filtering, while preserving val100 within roughly ±1 pp during OOD adaptation. TBLite does not improve under the same recipe, likely because its feedback is less directly action-linked.}
  \label{fig:env-only-cont}
\end{figure}

Figure~\ref{fig:env-only-cont} shows that env-only adaptation can improve a policy without verifier
rewards, but only when the interaction data provides useful prediction targets. On val100, where the checkpoint
is already competent, unfiltered env-only adaptation gives a $+3.8$~pp lift.
On harder OOD tasks, the policy more often enters bad interaction regimes: malformed tool calls, parse errors, and unproductive loops. In that regime, the environment loss can become a model of failure modes rather than useful task dynamics.
Filtering to clean tool-call trajectories removes this noise and yields sustained gains on PyTerm (+10.0 pp) and ITD (+5.2 pp).

Although TBLite starts at a similar pass rate to PyTerm, the same recipe degrades performance.
We suspect the difference is observation structure. PyTerm provides dense, action-linked feedback: code produces tracebacks, printed values, and file contents that often point directly to what should change next. TBLite often requires broader shell orchestration over less visible filesystem, configuration, and process state, so the observed terminal tokens are less directly tied to the action choices that would solve the task. Thus verifier-free env-only adaptation works best when
clean exploration exposes predictive, action-linked feedback. In those settings, a competent agent can continue improving from consequences alone: acting, observing what comes back, and updating only on the prediction loss.

\section{Related Work}
\label{sec:related}

ECHO builds on work showing that environment interaction contains useful supervision beyond sparse reward. Classical world-model methods learn dynamics models for planning, imagination, or search~\citep{schmidhuber1990making,ha2018worldmodels,hafner2020dreamer,hafner2021mastering,hafner2023dreamerv3,schrittwieser2020muzero,ye2021mastering}; recent embodied-agent work similarly uses world or action models to improve planning and control~\citep{wang2025world,ye2026world}. Auxiliary-prediction methods in model-free RL train agents to predict pixels, rewards, forward dynamics, future observations, or latent states to improve representations under sparse feedback~\citep{jaderberg2016reinforcement,pathak2017curiosity,schwarzer2021spr,kwon2024empirical}. ECHO follows this auxiliary-prediction view but in a multi-turn LM-agent setting: the targets are textual terminal observations already present in the transcript and already used as context for future actions.

For LM agents, the closest comparison is CWM~\citep{Copet2025CWMAO}, which trains a large model on observation--action trajectories from Python and Docker environments. Related recent work also uses agent experience or rich textual feedback to densify learning beyond scalar rewards~\citep{zhang2025agent,song2026expanding,hubotter2026reinforcement}. ECHO instead injects observation prediction directly into on-policy GRPO, requiring no separate corpus, world-modeling stage, feedback generator, dynamics model, or inference-time simulation. Some recent work also identifies the next-state/environment signal as discarded supervision. RLTF~\citep{song2026expanding} predicts judge-generated critiques as an auxiliary loss; OpenClaw-RL~\citep{wang2026openclaw} recovers next-state signals either as scalar process rewards via a judge, or as token-level distillation targets via judge-extracted hints. ECHO differs from both: we predict the raw environment-observation tokens directly via an auxiliary cross-entropy loss, with no judge, no critique, and no distillation — the supervision is the environment's literal response. It is also complementary to recent multi-turn agent RL systems such as SkyRL~\citep{Cao2025SkyRLAgentER}, SimpleTIR~\citep{xue2025simpletir}, DAPO~\citep{Yu2025DAPOAO}, and ArCHer~\citep{archer2024}: those methods stabilize the policy-gradient update, while ECHO adds a parallel supervised loss on the observation tokens.
\section{Conclusion}
\label{sec:conclusion}
We introduce ECHO, a simple way to turn the environment responses already present in agent rollouts into supervision. ECHO adds cross-entropy on terminal-output tokens to the same logits used for GRPO, requiring no extra rollouts, forward passes, data, or architectural changes. Across different starting policies and model sizes, ECHO improves over RL-only training, learns faster, matches much of the benefit of expert demonstrations, and roughly doubles TerminalBench-2.0 pass@1. 
Every action elicits an environment response, and every environment response is already in the agent rollout. This suggests a broader opportunity for agentic RL: between expert demonstrations and sparse outcome rewards there exists a dense training signal waiting to be used---the observable consequences of the agent's own actions.

\section*{Acknowledgments}
We thank MSR AI Frontiers for supporting this work. We are especially grateful to Ahmed Ghoneim for developing the data generation pipelines and datasets that supported these experiments. We greatly appreciate Vidhisha Balachandran, Sahaj Agarwal, Abhishek Goswami, and Mojan Javaheripi for helping build the agentic RL training and evaluation stacks in our internal codebase. We also thank John Langford for helpful early discussions.

\bibliographystyle{plainnat}
\bibliography{references}

\begin{thebibliography}{35}
\providecommand{\natexlab}[1]{#1}
\providecommand{\url}[1]{\texttt{#1}}
\expandafter\ifx\csname urlstyle\endcsname\relax
  \providecommand{\doi}[1]{doi: #1}\else
  \providecommand{\doi}{doi: \begingroup \urlstyle{rm}\Url}\fi

\bibitem[Cao et~al.(2025)Cao, Li, Zhao, Yuan, Hegde, Chen, Ruan, Griggs, Liu, Tang, Liaw, Moritz, Zaharia, Gonzalez, and Stoica]{Cao2025SkyRLAgentER}
Shiyi Cao, Dacheng Li, Fangzhou Zhao, Shuo Yuan, Sumanth~R. Hegde, Connor Chen, Charlie Ruan, Tyler Griggs, Shu Liu, Eric Tang, Richard Liaw, Philipp Moritz, Matei Zaharia, Joseph~E. Gonzalez, and Ion Stoica.
\newblock {SkyRL-Agent: Efficient RL Training for Multi-turn LLM Agent}, 2025.
\newblock URL \url{https://arxiv.org/abs/2511.16108}.

\bibitem[{FAIR CodeGen team} et~al.(2025){FAIR CodeGen team}, Copet, Carbonneaux, Cohen, Gehring, Kahn, Kossen, Kreuk, McMilin, Meyer, Wei, Zhang, Zheng, Armengol-Estap\'{e}, Bashiri, Beck, Chambon, Charnalia, Cummins, Decugis, Fisches, Fleuret, Gloeckle, Gu, Hassid, Haziza, Idrissi, Keller, Kindi, Leather, Maimon, Markosyan, Massa, Mazar\'{e}, Mella, Murray, Muzumdar, O'Hearn, Pagliardini, Pedchenko, Remez, Seeker, Selvi, Sultan, Wang, Wehrstedt, Yoran, Zhang, Cohen, Adi, and Synnaeve]{Copet2025CWMAO}
{FAIR CodeGen team}, Jade Copet, Quentin Carbonneaux, Gal Cohen, Jonas Gehring, Jacob Kahn, Jannik Kossen, Felix Kreuk, Emily McMilin, Michel Meyer, Yuxiang Wei, David Zhang, Kunhao Zheng, Jordi Armengol-Estap\'{e}, Pedram Bashiri, Maximilian Beck, Pierre Chambon, Abhishek Charnalia, Chris Cummins, Juliette Decugis, Zacharias~V. Fisches, Fran{\c c}ois Fleuret, Fabian Gloeckle, Alex Gu, Michael Hassid, Daniel Haziza, Badr~Youbi Idrissi, Christian Keller, Rahul Kindi, Hugh Leather, Gallil Maimon, Aram Markosyan, Francisco Massa, Pierre-Emmanuel Mazar\'{e}, Vegard Mella, Naila Murray, Keyur Muzumdar, Peter O'Hearn, Matteo Pagliardini, Dmitrii Pedchenko, Tal Remez, Volker Seeker, Marco Selvi, Oren Sultan, Sida Wang, Luca Wehrstedt, Ori Yoran, Lingming Zhang, Taco Cohen, Yossi Adi, and Gabriel Synnaeve.
\newblock {CWM: An Open-Weights LLM for Research on Code Generation with World Models}, 2025.
\newblock URL \url{https://arxiv.org/abs/2510.02387}.

\bibitem[Gandhi et~al.(2026)Gandhi, Garg, Goodman, and Papailiopoulos]{endlessterminals2026}
Kanishk Gandhi, Shivam Garg, Noah~D. Goodman, and Dimitris Papailiopoulos.
\newblock {{Endless Terminals}: Scaling {RL} Environments for Terminal Agents}, 2026.
\newblock URL \url{https://arxiv.org/abs/2601.16443}.

\bibitem[Ha and Schmidhuber(2018)]{ha2018worldmodels}
David Ha and J{\"u}rgen Schmidhuber.
\newblock Recurrent world models facilitate policy evolution.
\newblock In S.~Bengio, H.~Wallach, H.~Larochelle, K.~Grauman, N.~Cesa-Bianchi, and R.~Garnett, editors, \emph{Advances in Neural Information Processing Systems}, volume~31. Curran Associates, Inc., 2018.
\newblock URL \url{https://proceedings.neurips.cc/paper_files/paper/2018/file/2de5d16682c3c35007e4e92982f1a2ba-Paper.pdf}.

\bibitem[Hafner et~al.(2020)Hafner, Lillicrap, Ba, and Norouzi]{hafner2020dreamer}
Danijar Hafner, Timothy Lillicrap, Jimmy Ba, and Mohammad Norouzi.
\newblock Dream to control: Learning behaviors by latent imagination.
\newblock In \emph{International Conference on Learning Representations}, 2020.
\newblock URL \url{https://openreview.net/forum?id=S1lOTC4tDS}.

\bibitem[Hafner et~al.(2021)Hafner, Lillicrap, Norouzi, and Ba]{hafner2021mastering}
Danijar Hafner, Timothy Lillicrap, Mohammad Norouzi, and Jimmy Ba.
\newblock Mastering {A}tari with discrete world models.
\newblock In \emph{International Conference on Learning Representations}. OpenReview.net, 2021.
\newblock URL \url{https://openreview.net/pdf?id=0oabwyZbOu}.

\bibitem[Hafner et~al.(2025)Hafner, Pasukonis, Ba, and Lillicrap]{hafner2023dreamerv3}
Danijar Hafner, Jurgis Pasukonis, Jimmy Ba, and Timothy Lillicrap.
\newblock Mastering diverse control tasks through world models.
\newblock \emph{Nature}, 640\penalty0 (8059):\penalty0 647--653, April 2025.
\newblock \doi{10.1038/s41586-025-08744-2}.
\newblock URL \url{https://doi.org/10.1038/s41586-025-08744-2}.

\bibitem[{harbor-framework}(2025)]{harbor2025}
{harbor-framework}.
\newblock {Harbor}: A framework for evaluating and optimizing agents and models in container environments.
\newblock Software, August 2025.
\newblock URL \url{https://github.com/harbor-framework/harbor}.

\bibitem[{Harbor Framework Team}(2025)]{terminus2025}
{Harbor Framework Team}.
\newblock {Terminus-2}: Harbor's high-performance reference agent implementation.
\newblock Harbor documentation, 2025.
\newblock URL \url{https://www.harborframework.com/docs/agents/terminus-2}.
\newblock Accessed 2026-05-18.

\bibitem[H\"{u}botter et~al.(2026)H\"{u}botter, L\"{u}beck, Behric, Baumann, Bagatella, Marta, Hakimi, Shenfeld, Buening, Guestrin, and Krause]{hubotter2026reinforcement}
Jonas H\"{u}botter, Frederike L\"{u}beck, Lejs Behric, Anton Baumann, Marco Bagatella, Daniel Marta, Ido Hakimi, Idan Shenfeld, Thomas~Kleine Buening, Carlos Guestrin, and Andreas Krause.
\newblock {Reinforcement Learning via Self-Distillation}, 2026.
\newblock URL \url{https://arxiv.org/abs/2601.20802}.

\bibitem[Jaderberg et~al.(2017)Jaderberg, Mnih, Czarnecki, Schaul, Leibo, Silver, and Kavukcuoglu]{jaderberg2016reinforcement}
Max Jaderberg, Volodymyr Mnih, Wojciech~Marian Czarnecki, Tom Schaul, Joel~Z. Leibo, David Silver, and Koray Kavukcuoglu.
\newblock Reinforcement learning with unsupervised auxiliary tasks.
\newblock In \emph{International Conference on Learning Representations}, 2017.
\newblock URL \url{https://openreview.net/forum?id=SJ6yPD5xg}.

\bibitem[Kwon et~al.(2024)Kwon, Yang, Nowak, and Hanna]{kwon2024empirical}
Jeongyeol Kwon, Liu Yang, Robert Nowak, and Josiah Hanna.
\newblock An empirical study on the power of future prediction in partially observable environments, 2024.
\newblock URL \url{https://arxiv.org/abs/2402.07102}.

\bibitem[Merrill et~al.(2026)Merrill, Shaw, Carlini, Li, Raj, Bercovich, Shi, Shin, Walshe, Buchanan, Shen, Ye, Lin, Poulos, Wang, Nezhurina, Lu, Mastromichalakis, Xu, Chen, Liu, Zhang, Chen, Kashyap, Uslu, Li, Wu, Yan, Bian, Sharma, Sun, Dillmann, Anand, Lanpouthakoun, Koopah, Hu, Guha, Dreiman, Zhu, Krauth, Zhong, Muennighoff, Amanfu, Tan, Pimpalgaonkar, Aggarwal, Lin, Lan, Zhao, Liang, Wang, Wang, Zhou, Heineman, Liu, Trivedi, Yang, Lin, Shetty, Yang, Omi, Raoof, Li, Zhuo, Lin, Dai, Wang, Chai, Zhou, Wahdany, She, Hu, Dong, Zhu, Cui, Saiyed, Kolbeinsson, Rytting, Marten, Wang, Jitsev, Dimakis, Konwinski, and Schmidt]{terminalbench2025}
Mike~A Merrill, Alexander~Glenn Shaw, Nicholas Carlini, Boxuan Li, Harsh Raj, Ivan Bercovich, Lin Shi, Jeong~Yeon Shin, Thomas Walshe, E.~Kelly Buchanan, Junhong Shen, Guanghao Ye, Haowei Lin, Jason Poulos, Maoyu Wang, Marianna Nezhurina, Di~Lu, Orfeas~Menis Mastromichalakis, Zhiwei Xu, Zizhao Chen, Yue Liu, Robert Zhang, Leon~Liangyu Chen, Anurag Kashyap, Jan-Lucas Uslu, Jeffrey Li, Jianbo Wu, Minghao Yan, Song Bian, Vedang Sharma, Ke~Sun, Steven Dillmann, Akshay Anand, Andrew Lanpouthakoun, Bardia Koopah, Changran Hu, Etash~Kumar Guha, Gabriel H.~S. Dreiman, Jiacheng Zhu, Karl Krauth, Li~Zhong, Niklas Muennighoff, Robert~Kwesi Amanfu, Shangyin Tan, Shreyas Pimpalgaonkar, Tushar Aggarwal, Xiangning Lin, Xin Lan, Xuandong Zhao, Yiqing Liang, Yuanli Wang, Zilong Wang, Changzhi Zhou, David Heineman, Hange Liu, Harsh Trivedi, John Yang, Junhong Lin, Manish Shetty, Michael Yang, Nabil Omi, Negin Raoof, Shanda Li, Terry~Yue Zhuo, Wuwei Lin, Yiwei Dai, Yuxin Wang, Wenhao Chai, Shang Zhou, Dariush Wahdany, Ziyu She,
  Jiaming Hu, Zhikang Dong, Yuxuan Zhu, Sasha Cui, Ahson Saiyed, Arinbj{\"o}rn Kolbeinsson, Christopher~Michael Rytting, Ryan Marten, Yixin Wang, Jenia Jitsev, Alex Dimakis, Andy Konwinski, and Ludwig Schmidt.
\newblock Terminal-bench: Benchmarking agents on hard, realistic tasks in command line interfaces.
\newblock In \emph{The Fourteenth International Conference on Learning Representations}, 2026.
\newblock URL \url{https://openreview.net/forum?id=a7Qa4CcHak}.

\bibitem[{obiwan96}(2026)]{endlessterminalsdata2026}
{obiwan96}.
\newblock {Endless Terminals} dataset.
\newblock Hugging Face dataset, 2026.
\newblock URL \url{https://huggingface.co/datasets/obiwan96/endless-terminals}.
\newblock Associated with Gandhi et al., ``Endless Terminals: Scaling RL Environments for Terminal Agents''; accessed 2026-05-18.

\bibitem[{OpenThoughts-Agent Team}(2025{\natexlab{a}})]{openthinker2025}
{OpenThoughts-Agent Team}.
\newblock {OpenThinker-Agent-v1-SFT}.
\newblock Hugging Face model card, December 2025{\natexlab{a}}.
\newblock URL \url{https://huggingface.co/open-thoughts/OpenThinker-Agent-v1-SFT}.
\newblock Accessed 2026-05-18.

\bibitem[{OpenThoughts-Agent Team}(2025{\natexlab{b}})]{openthoughtsrl2025}
{OpenThoughts-Agent Team}.
\newblock {OpenThoughts-Agent-v1-RL}.
\newblock Hugging Face dataset, December 2025{\natexlab{b}}.
\newblock URL \url{https://huggingface.co/datasets/open-thoughts/OpenThoughts-Agent-v1-RL}.
\newblock Accessed 2026-05-18.

\bibitem[{OpenThoughts-Agent Team}(2025{\natexlab{c}})]{openthoughtstbdev2025}
{OpenThoughts-Agent Team}.
\newblock {OpenThoughts-TB-Dev}.
\newblock Hugging Face dataset, 2025{\natexlab{c}}.
\newblock URL \url{https://huggingface.co/datasets/open-thoughts/OpenThoughts-TB-dev}.
\newblock Accessed 2026-05-18.

\bibitem[{OpenThoughts-Agent team} et~al.(2026){OpenThoughts-Agent team}, {Snorkel AI}, and {Bespoke Labs}]{openthoughtstblite2025}
{OpenThoughts-Agent team}, {Snorkel AI}, and {Bespoke Labs}.
\newblock {OpenThoughts-TBLite}: A high-signal benchmark for iterating on terminal agents.
\newblock Hugging Face dataset, February 2026.
\newblock URL \url{https://huggingface.co/datasets/open-thoughts/OpenThoughts-TBLite}.
\newblock Accessed 2026-05-18.

\bibitem[Ouyang et~al.(2022)Ouyang, Wu, Jiang, Almeida, Wainwright, Mishkin, Zhang, Agarwal, Slama, Ray, Schulman, Hilton, Kelton, Miller, Simens, Askell, Welinder, Christiano, Leike, and Lowe]{rlhf-instruct2022}
Long Ouyang, Jeff Wu, Xu~Jiang, Diogo Almeida, Carroll~L. Wainwright, Pamela Mishkin, Chong Zhang, Sandhini Agarwal, Katarina Slama, Alex Ray, John Schulman, Jacob Hilton, Fraser Kelton, Luke~E. Miller, Maddie Simens, Amanda Askell, Peter Welinder, Paul~Francis Christiano, Jan Leike, and Ryan~J. Lowe.
\newblock Training language models to follow instructions with human feedback.
\newblock In S.~Koyejo, S.~Mohamed, A.~Agarwal, D.~Belgrave, K.~Cho, and A.~Oh, editors, \emph{Advances in Neural Information Processing Systems}, volume~35, pages 27730--27744. Curran Associates, Inc., 2022.
\newblock URL \url{https://proceedings.neurips.cc/paper_files/paper/2022/file/b1efde53be364a73914f58805a001731-Paper-Conference.pdf}.

\bibitem[Patel(2023)]{sutskever2023dwarkesh}
Dwarkesh Patel.
\newblock Ilya sutskever (openai chief scientist) -- why next-token prediction could surpass human intelligence.
\newblock Interview by Dwarkesh Patel, Dwarkesh Podcast, March 2023.
\newblock URL \url{https://www.dwarkesh.com/p/ilya-sutskever}.
\newblock Transcript, accessed 2026-05-18.

\bibitem[Pathak et~al.(2017)Pathak, Agrawal, Efros, and Darrell]{pathak2017curiosity}
Deepak Pathak, Pulkit Agrawal, Alexei~A Efros, and Trevor Darrell.
\newblock Curiosity-driven exploration by self-supervised prediction.
\newblock In \emph{International conference on machine learning}, pages 2778--2787. PMLR, 2017.
\newblock URL \url{https://proceedings.mlr.press/v70/pathak17a.html}.

\bibitem[Schmidhuber(1990)]{schmidhuber1990making}
J{\"u}rgen Schmidhuber.
\newblock Making the world differentiable: On using self-supervised fully recurrent neural networks for dynamic reinforcement learning and planning in non-stationary environments.
\newblock Technical Report FKI-126-90, Technische Universit{\"a}t M{\"u}nchen, 1990.
\newblock URL \url{https://people.idsia.ch/~juergen/FKI-126-90_%28revised%29bw_ocr.pdf}.

\bibitem[Schrittwieser et~al.(2020)Schrittwieser, Antonoglou, Hubert, Simonyan, Sifre, Schmitt, Guez, Lockhart, Hassabis, Graepel, Lillicrap, and Silver]{schrittwieser2020muzero}
Julian Schrittwieser, Ioannis Antonoglou, Thomas Hubert, Karen Simonyan, Laurent Sifre, Simon Schmitt, Arthur Guez, Edward Lockhart, Demis Hassabis, Thore Graepel, Timothy Lillicrap, and David Silver.
\newblock Mastering {A}tari, {Go}, chess and shogi by planning with a learned model.
\newblock \emph{Nature}, 588\penalty0 (7839):\penalty0 604--609, 2020.
\newblock \doi{10.1038/s41586-020-03051-4}.
\newblock URL \url{https://doi.org/10.1038/s41586-020-03051-4}.

\bibitem[Schwarzer et~al.(2021)Schwarzer, Anand, Goel, Hjelm, Courville, and Bachman]{schwarzer2021spr}
Max Schwarzer, Ankesh Anand, Rishab Goel, R~Devon Hjelm, Aaron Courville, and Philip Bachman.
\newblock Data-efficient reinforcement learning with self-predictive representations.
\newblock In \emph{International Conference on Learning Representations}, 2021.
\newblock URL \url{https://arxiv.org/abs/2007.05929}.

\bibitem[Shao et~al.(2024)Shao, Wang, Zhu, Xu, Song, Bi, Zhang, Zhang, Li, Wu, and Guo]{Shao2024DeepSeekMathPT}
Zhihong Shao, Peiyi Wang, Qihao Zhu, Runxin Xu, Junxiao Song, Xiao Bi, Haowei Zhang, Mingchuan Zhang, Y.~K. Li, Y.~Wu, and Daya Guo.
\newblock {{DeepSeekMath}: Pushing the Limits of Mathematical Reasoning in Open Language Models}, 2024.
\newblock URL \url{https://arxiv.org/abs/2402.03300}.

\bibitem[Song et~al.(2026)Song, Chen, Tajwar, Munos, Pathak, Bagnell, Singh, and Zanette]{song2026expanding}
Yuda Song, Lili Chen, Fahim Tajwar, Remi Munos, Deepak Pathak, J.~Andrew Bagnell, Aarti Singh, and Andrea Zanette.
\newblock Expanding the capabilities of reinforcement learning via text feedback, 2026.
\newblock URL \url{https://arxiv.org/abs/2602.02482}.

\bibitem[Wang et~al.(2025)Wang, Fei, Cheng, Zhang, Cai, Fu, and Qiu]{wang2025world}
Siyin Wang, Zhaoye Fei, Qinyuan Cheng, Shiduo Zhang, Panpan Cai, Jinlan Fu, and Xipeng Qiu.
\newblock World modeling makes a better planner: Dual preference optimization for embodied task planning.
\newblock In \emph{Proceedings of the 63rd Annual Meeting of the Association for Computational Linguistics (Volume 1: Long Papers)}, pages 21518--21537, 2025.
\newblock URL \url{https://aclanthology.org/2025.acl-long.1044/}.

\bibitem[Wang et~al.(2026)Wang, Chen, Jin, Wang, and Yang]{wang2026openclaw}
Yinjie Wang, Xuyang Chen, Xiaolong Jin, Mengdi Wang, and Ling Yang.
\newblock {OpenClaw-RL: Train Any Agent Simply by Talking}, 2026.
\newblock URL \url{https://arxiv.org/abs/2603.10165}.

\bibitem[Xue et~al.(2026)Xue, Zheng, Liu, Li, Zheng, MA, and An]{xue2025simpletir}
Zhenghai Xue, Longtao Zheng, Qian Liu, Yingru Li, Xiaosen Zheng, Zejun MA, and Bo~An.
\newblock Simple{TIR}: End-to-end reinforcement learning for multi-turn tool-integrated reasoning.
\newblock In \emph{The Fourteenth International Conference on Learning Representations}, 2026.
\newblock URL \url{https://openreview.net/forum?id=EplNy91Xqh}.

\bibitem[Yang et~al.(2025)Yang, Li, Yang, Zhang, Hui, Zheng, Yu, Gao, Huang, Lv, Zheng, Liu, Zhou, Huang, Hu, Ge, Wei, Lin, Tang, Yang, Tu, Zhang, Yang, Yang, Zhou, Zhou, Lin, Dang, Bao, Yang, Yu, Deng, Li, Xue, Li, Zhang, Wang, Zhu, Men, Gao, Liu, Luo, Li, Tang, Yin, Ren, Wang, Zhang, Ren, Fan, Su, Zhang, Zhang, Wan, Liu, Wang, Cui, Zhang, Zhou, and Qiu]{qwen3-2025}
An~Yang, Anfeng Li, Baosong Yang, Beichen Zhang, Binyuan Hui, Bo~Zheng, Bowen Yu, Chang Gao, Chengen Huang, Chenxu Lv, Chujie Zheng, Dayiheng Liu, Fan Zhou, Fei Huang, Feng Hu, Hao Ge, Haoran Wei, Huan Lin, Jialong Tang, Jian Yang, Jianhong Tu, Jianwei Zhang, Jianxin Yang, Jiaxi Yang, Jing Zhou, Jingren Zhou, Junyang Lin, Kai Dang, Keqin Bao, Kexin Yang, Le~Yu, Lianghao Deng, Mei Li, Mingfeng Xue, Mingze Li, Pei Zhang, Peng Wang, Qin Zhu, Rui Men, Ruize Gao, Shixuan Liu, Shuang Luo, Tianhao Li, Tianyi Tang, Wenbiao Yin, Xingzhang Ren, Xinyu Wang, Xinyu Zhang, Xuancheng Ren, Yang Fan, Yang Su, Yichang Zhang, Yinger Zhang, Yu~Wan, Yuqiong Liu, Zekun Wang, Zeyu Cui, Zhenru Zhang, Zhipeng Zhou, and Zihan Qiu.
\newblock Qwen3 technical report, 2025.
\newblock URL \url{https://arxiv.org/abs/2505.09388}.

\bibitem[Ye et~al.(2026)Ye, Ge, Zheng, Gao, Yu, Kurian, Indupuru, Tan, Zhu, Xiang, Malik, Lee, Liang, Ranawaka, Gu, Xu, Wang, Hu, Narayan, Bjorck, Wang, Kim, Niu, Zheng, Xie, Wu, Wang, Julian, Xu, Du, Chebotar, Reed, Kautz, Zhu, Fan, and Jang]{ye2026world}
Seonghyeon Ye, Yunhao Ge, Kaiyuan Zheng, Shenyuan Gao, Sihyun Yu, George Kurian, Suneel Indupuru, You~Liang Tan, Chuning Zhu, Jiannan Xiang, Ayaan Malik, Kyungmin Lee, William Liang, Nadun Ranawaka, Jiasheng Gu, Yinzhen Xu, Guanzhi Wang, Fengyuan Hu, Avnish Narayan, Johan Bjorck, Jing Wang, Gwanghyun Kim, Dantong Niu, Ruijie Zheng, Yuqi Xie, Jimmy Wu, Qi~Wang, Ryan Julian, Danfei Xu, Yilun Du, Yevgen Chebotar, Scott Reed, Jan Kautz, Yuke Zhu, Linxi~"Jim" Fan, and Joel Jang.
\newblock World action models are zero-shot policies, 2026.
\newblock URL \url{https://arxiv.org/abs/2602.15922}.

\bibitem[Ye et~al.(2021)Ye, Liu, Kurutach, Abbeel, and Gao]{ye2021mastering}
Weirui Ye, Shaohuai Liu, Thanard Kurutach, Pieter Abbeel, and Yang Gao.
\newblock Mastering {A}tari games with limited data.
\newblock \emph{Advances in neural information processing systems}, 34:\penalty0 25476--25488, 2021.
\newblock URL \url{https://proceedings.neurips.cc/paper_files/paper/2021/file/d5eca8dc3820cad9fe56a3bafda65ca1-Paper.pdf}.

\bibitem[Yu et~al.(2025)Yu, Zhang, Zhu, Yuan, Zuo, YuYue, Dai, Fan, Liu, Liu, Liu, Liu, Lin, Lin, Ma, Sheng, Tong, Zhang, Zhang, Zhang, Zhang, Zhu, Zhu, Chen, Chen, Wang, Yu, Song, Wei, Zhou, Liu, Ma, Zhang, Yan, Wu, and Wang]{Yu2025DAPOAO}
Qiying Yu, Zheng Zhang, Ruofei Zhu, Yufeng Yuan, Xiaochen Zuo, YuYue, Weinan Dai, Tiantian Fan, Gaohong Liu, Juncai Liu, LingJun Liu, Xin Liu, Haibin Lin, Zhiqi Lin, Bole Ma, Guangming Sheng, Yuxuan Tong, Chi Zhang, Mofan Zhang, Ru~Zhang, Wang Zhang, Hang Zhu, Jinhua Zhu, Jiaze Chen, Jiangjie Chen, Chengyi Wang, Hongli Yu, Yuxuan Song, Xiangpeng Wei, Hao Zhou, Jingjing Liu, Wei-Ying Ma, Ya-Qin Zhang, Lin Yan, Yonghui Wu, and Mingxuan Wang.
\newblock {DAPO}: An open-source {LLM} reinforcement learning system at scale.
\newblock In \emph{The Thirty-ninth Annual Conference on Neural Information Processing Systems}, 2025.
\newblock URL \url{https://openreview.net/forum?id=2a36EMSSTp}.

\bibitem[Zhang et~al.(2025)Zhang, Chen, Liu, Xue, Liao, Liu, Wang, Ning, Chen, Fu, Xie, Sun, Gou, Qi, Meng, Yang, Zhang, Li, Shah, Huynh, Li, Yang, Cao, Jang, Zhou, Zhu, Sun, Weston, Su, and Wu]{zhang2025agent}
Kai Zhang, Xiangchao Chen, Bo~Liu, Tianci Xue, Zeyi Liao, Zhihan Liu, Xiyao Wang, Yuting Ning, Zhaorun Chen, Xiaohan Fu, Jian Xie, Yuxuan Sun, Boyu Gou, Qi~Qi, Zihang Meng, Jianwei Yang, Ning Zhang, Xian Li, Ashish Shah, Dat Huynh, Hengduo Li, Zi~Yang, Sara Cao, Lawrence Jang, Shuyan Zhou, Jiacheng Zhu, Huan Sun, Jason Weston, Yu~Su, and Yifan Wu.
\newblock Agent learning via early experience, 2025.
\newblock URL \url{https://arxiv.org/abs/2510.08558}.

\bibitem[Zhou et~al.(2024)Zhou, Zanette, Pan, Levine, and Kumar]{archer2024}
Yifei Zhou, Andrea Zanette, Jiayi Pan, Sergey Levine, and Aviral Kumar.
\newblock {ArCHer}: Training language model agents via hierarchical multi-turn {RL}, 2024.
\newblock URL \url{https://arxiv.org/abs/2402.19446}.

\end{thebibliography}

\appendix
\section*{Appendix}
\section{Environment-Token Cross-Entropy Trajectories}
\label{app:ce-traj}

\begin{figure}[h]
  \centering
  \includegraphics[width=0.7\linewidth]{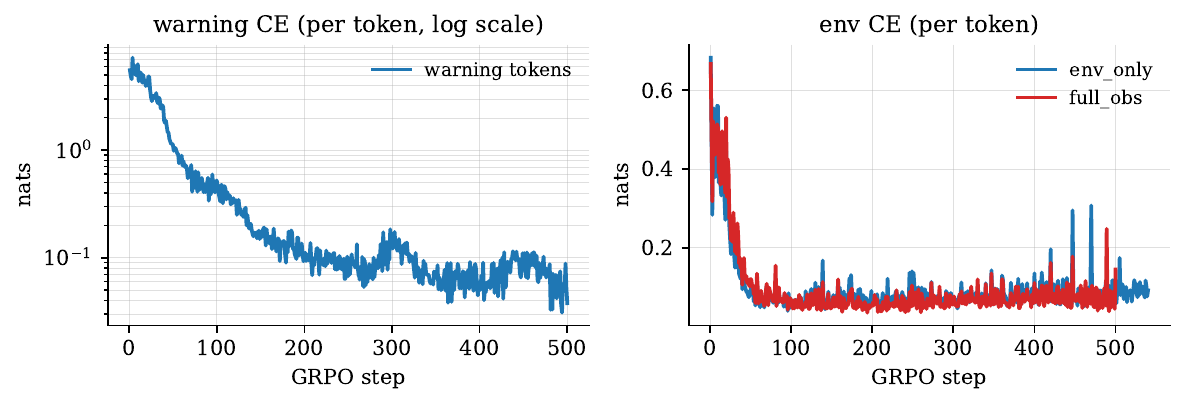}
  \captionsetup{font=small, labelfont={small,bf}}
  \caption{\textbf{Per-token environment cross-entropy by target type.} Warning CE drops to near-zero within ${\sim}60$ steps; env CE plateaus at $0.05$--$0.10$ nats and provides sustained gradient throughout training.}
  \label{fig:ce-trajectories}
\end{figure}

Figure~\ref{fig:ce-trajectories} shows per-token environment cross-entropy on warning tokens and on terminal-output (env) tokens over training. Warning CE drops from ${\sim}5.6$ nats to ${<}0.05$ nats by step 60---the model memorizes warning structure quickly. Env CE plateaus at $0.05$--$0.10$ nats, the irreducible entropy of real terminal output (variable filenames, byte counts, error formats). A constant $\lambda$ schedule therefore auto-anneals the warning gradient while preserving the env gradient indefinitely; this motivates the $\mathcal{O}'\!\!=\!\text{env\_only}$ choice in Section~\ref{sec:method-target}.

\section{Hyperparameters and Reproducibility}
\label{app:hyperparams}

\begin{itemize}
  \item Optimizer: AdamW, $\beta_1=0.9$, $\beta_2=0.95$, weight decay $0.01$.
  \item Learning rate: $1\times10^{-6}$ constant (no warmup or decay), gradient clip $0.2$.
  \item GRPO: $n=16$ rollouts per prompt, batch size 16, no KL penalty, prompt-level advantage normalization, $\epsilon_{\text{lo}}=0.2$, $\epsilon_{\text{hi}}=0.28$.
  \item Sampling: training temperature $0.8$; evaluation temperature $0.6$.
  \item Environment-prediction loss: $\lambda \in \{0.02,\,0.05\}$ (SFT vs.\ base); $\mathcal{O}'$ = terminal-output (env) tokens; per-sequence normalization by total observation length.
  \item Run length: 500--1000 GRPO steps on 8 GPUs (A100/B200 mix), ${\sim}24$--$48$~h wallclock per run.
\end{itemize}

\paragraph{Internal-Eval Reproducibility.} 8 rollouts/task, sampling temperature 0.6, 16-turn budget. Per-task pass-rate variance at $n=8$ is ${\pm}0.05$; mean-of-100-task variance is approximately ${\pm}0.025$ (Wilson interval). We treat trends across $\geq 3$ consecutive eval points as more informative than single-checkpoint deltas.

\paragraph{TerminalBench-2.0 Reproducibility.} Terminus-2 reference agent at temperature 0.6, $n_{\text{attempts}}=5$, 1200~s agent and verifier timeouts, sampling seed 42. Standard error on pass@1 at $n=5$ is ${\sim}1.5$~pp.

\section{Expert-SFT Gap}
\label{app:sftgap-full}

Table~\ref{tab:sftgap-full} reports the absolute pass-rates for the three Qwen3-8B configurations compared in \S\ref{sec:sftgap}, together with the per-metric SFT gap (SFT$+$GRPO minus GRPO), the lift from ECHO (ECHO - GRPO), and the fraction of the SFT gap recovered by ECHO without using expert demonstrations. Values are in pass-rate units (\%); internal columns (val100, ITD, TBLite) are mean pass-rate, TB2 columns are pass@$k$.

\begin{table}[h!]
\centering
\small
\setlength{\tabcolsep}{4pt}
\begin{tabular}{lcccccc}
\toprule
\textbf{Configuration} & \textbf{val100} & \textbf{ITD} & \textbf{TBLite} & \textbf{TB2 p@1} & \textbf{TB2 p@3} & \textbf{TB2 p@5} \\
\midrule
\rowcolor{rowbase} Qwen3-8B GRPO              & 54.94 & 16.22 &  9.47 & 2.70  & 6.74  &  8.99 \\
\rowcolor{rowecho} Qwen3-8B ECHO   & 63.66 & 18.89 & 11.39 & 5.17  & 10.45 & 13.48 \\
\rowcolor{rowrl}   Qwen3-8B SFT+GRPO    & 63.52 & 18.79 & 11.63 & 7.64  & 14.38 & 17.98 \\
\midrule
SFT gap (SFT$+$GRPO $-$ GRPO)     &  8.58 &  2.57 &  2.16 & 4.94 &  7.64 &  8.99 \\
ECHO lift (ECHO $-$ GRPO)  &  8.72 &  2.67 &  1.92 & 2.47 &  3.71 &  4.49 \\
\textbf{\% SFT gap closed by ECHO} & \textbf{101.6\%} & \textbf{103.9\%} & \textbf{88.9\%} & \textbf{50.0\%} & \textbf{48.6\%} & \textbf{50.0\%} \\
\bottomrule
\end{tabular}
\captionsetup{font=small, labelfont={small,bf}}
\caption{\textbf{Full breakdown of the expert-SFT gap closed by ECHO on Qwen3-8B base.} Rows are absolute pass-rates for the three matched configurations, the SFT initialization gap, the lift from adding ECHO without expert SFT, and the fraction of that gap closed.}
\label{tab:sftgap-full}
\end{table}
\section{OT-SFT Training Curves}
\label{app:otsft-curves}

\begin{figure}[h]
  \centering
  \includegraphics[width=\linewidth]{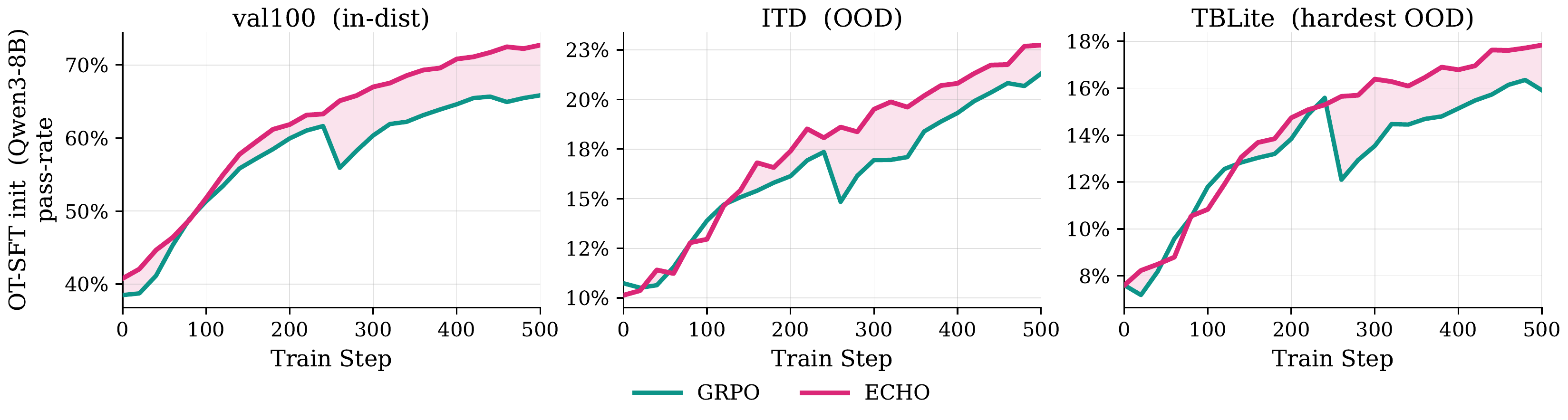}
  \captionsetup{font=small, labelfont={small,bf}}
  \caption{\textbf{OT-SFT-init Qwen3-8B training curves.}}
  \label{fig:curves-otsft}
\end{figure}

Figure~\ref{fig:curves-otsft} shows training curves for OpenThinker-Agent-v1-SFT Qwen3-8B. The curves follow a similar trend as Qwen3-8B and Qwen3-14B with ECHO quickly surpassing GRPO and remaining consistently ahead during training.

\end{document}